WLC-Net: a robust and fast deep-learning wood-leaf classification method

Hanlong Li, Pei Wang, Yuhan Wu, Jing Ren, Yuhang Gao, Lingyun Zhang, Mingtai Zhang, Wenxin Chen

Abstract: Wood-leaf classification is an essential and fundamental prerequisite in the analysis and estimation of forest attributes from terrestrial laser scanning (TLS) point clouds, including critical measurements such as diameter at breast height (DBH), above-ground biomass (AGB), wood volume. To address this, we introduce the Wood-Leaf Classification Network (WLC-Net), a deep learning model derived from PointNet++, designed to differentiate between wood and leaf points within tree point clouds. WLC-Net enhances classification accuracy, completeness, and speed by incorporating linearity as an inherent feature, refining the input-output framework, and optimizing the centroid sampling technique. WLC-Net was trained and assessed using three distinct tree species datasets, comprising a total of 102 individual tree point clouds: 21 Chinese ash trees, 21 willow trees, and 60 tropical trees. For comparative evaluation, five alternative methods, including PointNet++, DGCNN, Krishna Moorthy's method, LeWoS, and Sun's method, were also applied to these datasets. The classification accuracy of all six methods was quantified using three metrics: overall accuracy (OA), mean Intersection over Union (mIoU), and F1-score. Across all three datasets, WLC-Net demonstrated superior performance, achieving OA scores of 0.9778, 0.9712, and 0.9508; mIoU scores of 0.9761, 0.9693, and 0.9141; and F1-scores of 0.8628, 0.7938, and 0.9019, respectively. The time costs of WLC-Net were also recorded to evaluate the efficiency. The average processing time was 102.74s per million points for WLC-Net. In terms of visual inspect, accuracy evaluation and efficiency evaluation, the results suggest that WLC-Net presents a promising approach for wood-leaf classification, distinguished by its high accuracy. WLC-Net has demonstrated certain advantages in wood-leaf classification. In addition, WLC-Net also exhibits strong applicability across various tree point clouds and holds promise for further optimization.

# 1. Introduction

Terrestrial laser scanner emits laser beams and measures their return time to accurately calculate the positions of reflection points on the object. The detailed three-dimensional insights are very important and helpful for understanding forest ecosystems and enhancing strategic planning in forest management. In the past decades, TLS technique is increasingly popular in forestry due to its exceptional precision and efficiency which can promote non-destructive measurements of the related forest research, such as DBH, AGB, canopy structure, tree skeleton extraction etc.

Within the realm of forestry studies, wood-leaf classification stands as an essential and fundamental key step. For example, the extraction of leaf information from tree point cloud can significantly contribute to the study of tree physiology, the evaluation of tree health, and the comprehension of carbon uptake dynamics within forest ecosystems. While the analysis of woody structures derived from point cloud data can facilitate the estimation of DBH, tree volume and AGB through the application of quantitative structural models (Calders et al., 2015; Kükenbrink et al. 2021). The delineation of tree point clouds into wood and leaf components is thus instrumental in enabling precise analysis and the development of innovative applications, encompassing the exchange of mass and energy, as well as the partitioning of net primary production across various components (Hosoi et al., 2010; Kong et al., 2016; Olsoy et al., 2016). However, the inherent irregularity, complexity, and unstructured nature of three-dimensional forestry point clouds, also present significant challenges to the task of wood-leaf classification. There has been a burgeoning interest in the wood-leaf classification based on tree point cloud. And current approaches can be broadly categorized into three principal methodologies: traditional method, machine learning method, and deep learning method.

Basically, traditional methods for wood-leaf classification from tree point clouds have predominantly relied on geometric or radiometric features, or a combination of both. The geometric information of tree point cloud is frequently utilized as the primary classification criterion due to the more pronounced and structured nature of trunk and branch structures, in contrast to the more dispersed and random arrangement of leaves (Sun et al., 2021). For example, the LeWoS, an automatic MATLAB-based tool designed for wood-leaf classification, employs classification outcomes for tree modeling and subsequently benchmarks these results against those derived from manually classified wood points (Wang et al., 2019). Additionally, Sun et al. proposed an automated three-step classification approach that integrates the intensity data, neighborhood relationships, and point cloud density to facilitate fast wood-leaf classification (Sun et al., 2021). These methods are often contingent on the selection of specific parameters, which can significantly influence the classification outcomes. Furthermore, the applicability of these methods can be constrained when fundamental data, such as intensity information, is absent. For instance, Sun's three-step method is contingent upon the availability of intensity data for its application.

While these methods have been proven effective in wood--leaf classification, they are with the requirement for labor-intensive manual annotation or specific user-defined parameters(Dong et al., 2020). This requirement not only renders the process time-consuming but also susceptible to human error, which can compromise the reliability and scalability of the methods. Additionally, the inherent complexity of forestry point clouds, with their intricate branching structures and varying densities, presents further challenges to achieve accurate and automated classification(Jiang et al., 2023). Therefore, there is an urgent for more robust, automated solutions capable of managing the intricacies of forestry point clouds while reducing the reliance on manual intervention.

Machine-learning also have been increasingly applied to the wood-leaf classification of tree point clouds. These approaches generally encompass two pivotal stages: the extraction of distinctive features to act as classifiers, and the utilization of machine learning techniques to categorize each point into either wood or leaf points. Both supervised and unsupervised machine learning algorithms have been leveraged in this context, such as Support Vector Machines (SVM)(Yun et al., 2016; Liu et al., 2020), Random Forests (RF) (Zhu et al., 2018), Gauss Mixture Models (GMM)(Ma et al., 2015) and Density-Based Spatial Clustering of Applications with Noise

(DBSCAN)(Ferrara et al., 2018). Additionally, hybrid techniques have been integrated into classical machine learning methodologies for wood-leaf classification. For instance, a framework that integrates unsupervised classification of geometric features with shortest path analysis has achieved an average accuracy of 89% with field data and 83% with simulated data (Vicari et al., 2019). Similarly, a sophisticated supervised learning approach, which combines geometrical features defined by radially bounded nearest neighbors across multiple spatial scales, has yielded an overall average accuracy of 89.75% on field data from tropical and deciduous forests, and 82.1% on simulated point clouds(Krishna Moorthy et al., 2020). Despite these methods' impressive performance on test datasets, they continue to confront the challenge posed by the structural complexity inherent in tree point cloud.

Beyond the realm of machine learning, the emergence of deep learning has introduced a transformative approach to wood-leaf classification. Proficient at discerning complex patterns within high-dimensional data, deep learning algorithms effectively capture the subtle, non-linear interplays between inputs and outputs (Fei et al. 2017; Wickramasinghe et al. 2022), promising significant enhancements to classification accuracy. Early deep learning forays involved converting point clouds into projections or voxels, exemplified by MVCNN、View-GCN and VoxNet. However, these transformations techniques risked the loss of explicit information (Su et al., 2015; Li et al., 2020; Maturana and Scherer, 2015; Wu et al., 2019) and demanded substantial computational resources and advanced hardware. A paradigm shift occurred with the advent of PointNet (Qi et al., 2017), an algorithm engineered specifically for direct processing of raw point cloud data. Despite its pioneering role, PointNet faced limitations capturing the local structures and leveraging multi-scale features. To address these shortcomings, subsequent methods, such as PointNet++(Qi et al., 2017)、MO-Net、DGCNN(Wang et al., 2021)、PointCNN(Li et al., 2018)、D-FCN(Wen et al., 2020) were developed. Additionally, specialized deep learning frameworks for wood-leaf classification, including MDC-Net(Dai et al., 2023) and FWCNN(Wu et al., 2020), have been progressively introduced. While these advanced methods offer the promising avenues for enhancing accuracy in wood-leaf classification, the full integration of deep learning within forestry classification is still evolving. The field necessitates ongoing research to fully harness the capabilities of deep learning and achieve its potential.

In this study, we proposed the specific WLC-Net, a novel deep learning approach tailored for wood-leaf classification in tree point clouds. WLC-Net is constructed upon the foundational architecture of PointNet++, with three key innovations: introduce linearity as a prior feature, automatically output complete prediction results, and optimize the centroid sampling technique. All the three methods help to achieve improvements in both the accuracy and efficiency of wood-leaf classification. The incorporation of linearity as an inherent feature, an automated mechanism for yielding comprehensive prediction outcomes, and an enhanced centroid sampling technique. Collectively, these innovations contribute to significant enhancements in both the precision and efficiency of the wood-leaf classification process.

## 2. Materials and Methods

### 2.1. Equipment and Data

The data were collected using the RIEGL VZ-400 scanner, a sophisticated laser measurement system manufactured by RIEGL Laser Measurement Systems GmbH, located in Horn, Austria). This system is distinguished for its extensive field of view, which covers a full 360° horizontally and spans 100° vertically from +60° to -40°, with angular resolutions of less than 0.0005°. Additionally, the scanner's proficiency in generating up to 300,000 points per second significantly enhances the density and precision of resultant point clouds. The comprehensive characteristics of the scanner are listed in Table 1.

**Table 1.** The characteristics of RIEGL VZ-400 scanner.

| Technical Parameters | |
|---|---|
| Maximum scanning distance | 600m (Natural object reflectivity ≥90%) |
| Vertical Scan Angle Range | Total 100° (+60°/-40°) |
| Horizontal Scan Angle Range | Max 360° |
| Accuracy | 5mm |
| Scan Speed | 3 lines/sec to 120 lines/sec (Vertical) |
| | 0°/sec to 60°/sec (Horizontal) |
| Laser Pulse Repetition Rate | 100 kHz (Long Range Mode) |
| | 300kHz (High Speed Mode) |
| Angular Resolution | Better than 0.0005° |

The experiment was carried out in July 2023 at the Dongsheng Bajia Suburban Park, situated in the Haidian District of Beijing. The study focused on two tree species: Chinese ash (*Fraxinus chinensis Roxb.*) and willow (*Salix matsudana Koidz.*). The dataset included a total of 42 trees, specifically consisting of 21 Chinese ash trees and 21 willow trees. The scanning angular resolution is set 0.1 degrees. The tree heights spanned from 6.68 meters to 18.09 meters, and their diameters at breast height (DBH) ranged from 10.3 cm to 43.7 cm. Each tree point cloud was assigned a unique numerical identifier and demonstrated in Figure 1 and Figure2, respectively.

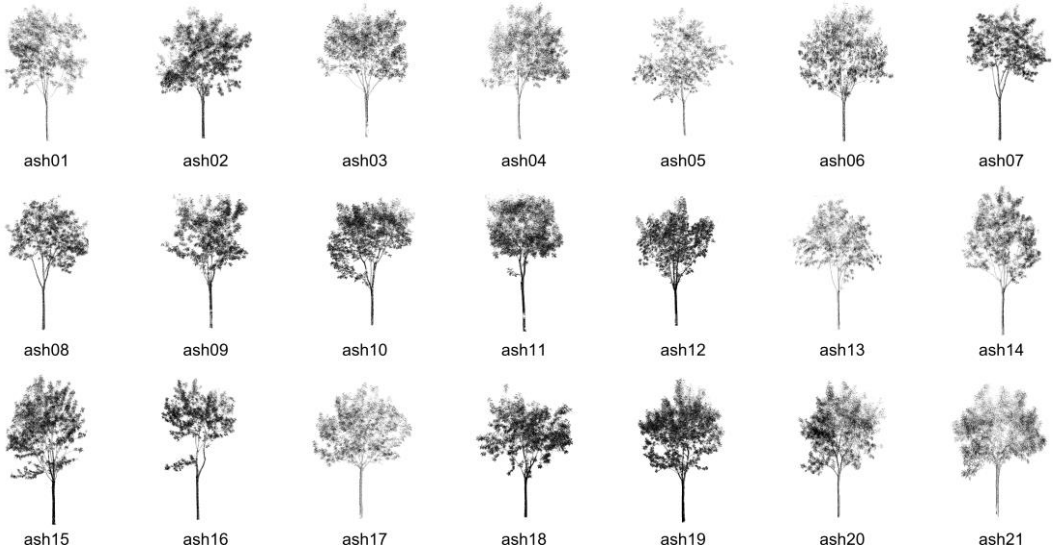

**Figure 1.** Extracted 21 Chinese ash point clouds and their numerical identifier.

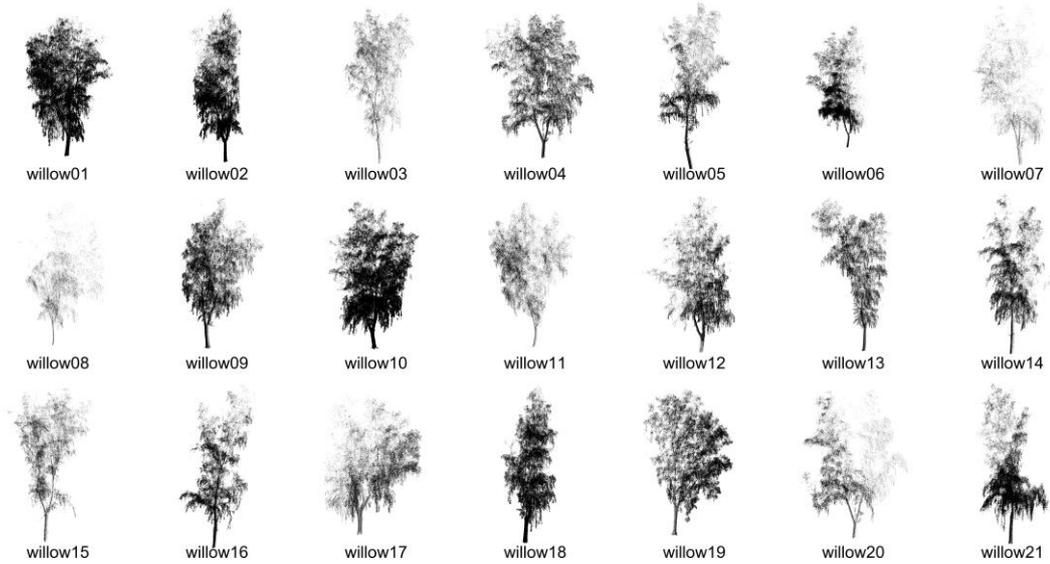

**Figure 2.** Extracted 21 Willow point clouds and their numerical identifier.

All the trees point clouds are manual extracted and meticulously classified into wood and leaf categories to establish a benchmark for wood-leaf classification using CloudCompare, which is an open-source software licensed under the GNU General Public License (GPL). The standard classification for an exemplar tree, designated as 'ash07', is illustrated in Figure 3, where wood points are represented in brown and leaf points in green. The manual extraction and classification effort totaled costed about 140 hours. All processing procedures were conducted on a laptop equipped with an NVIDIA GeForce RTX 3080, 16GB of memory, and an Intel Core i7-11800H CPU.

Furthermore, an open-source dataset (Wang et al., 2020) was used to verify the effectiveness of WLC-Net. This dataset encompasses 61 tropical trees across 15 distinct species. Tree heights in the dataset ranged from 8.7 to 53.6 meters, with an average height of 33.7 ± 12.4 meters. Similarly, DBH ranged from 10.8 to 186.6 centimeters, averaging at 58.4 ± 41.3 centimeters.

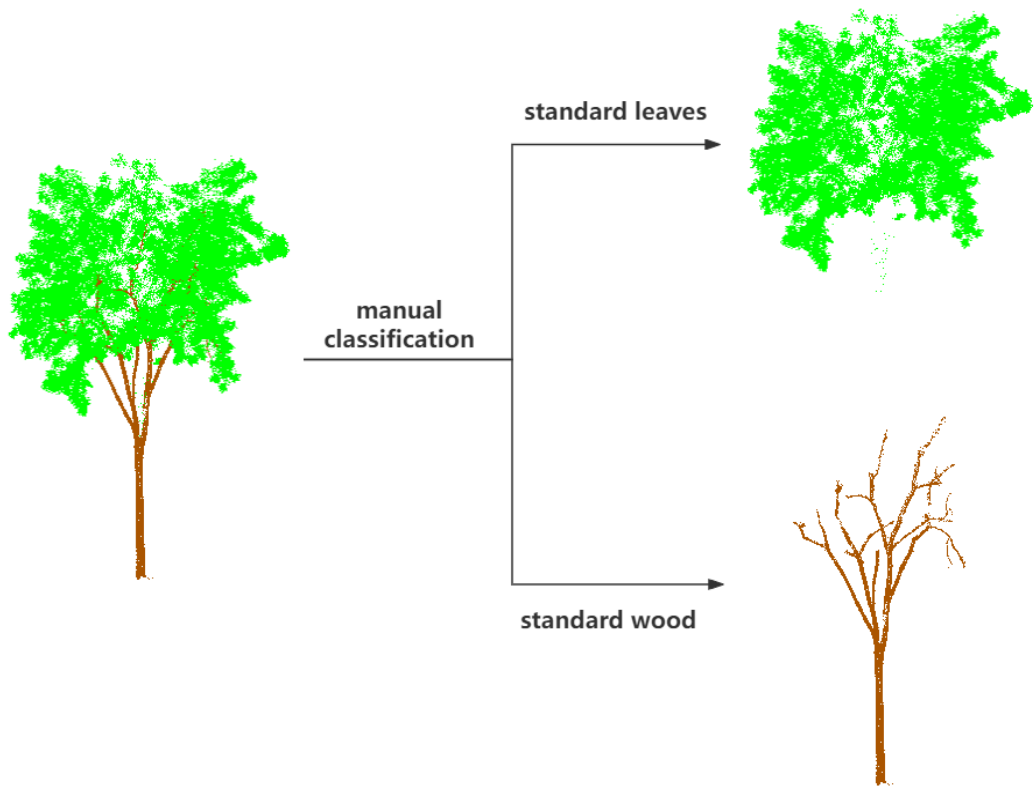

**Figure 3.** The manual standard classification result of ash 7. Brown: wood points; green: leaf points.

## 2.2. Overview of Method

PointNet++, a framework renowned in the deep learning domain, has exhibited robust performance across a spectrum of evaluations. Building upon the capabilities of PointNet++, we introduce WLC-Net to classify the tree point cloud into wood and leaf points. As depicted in Figure 4, the architecture of WLC-Net ingests the initial 3D coordinates and preliminary features to generate semantic labels for each point, thereby enhancing the granularity of classification outputs.

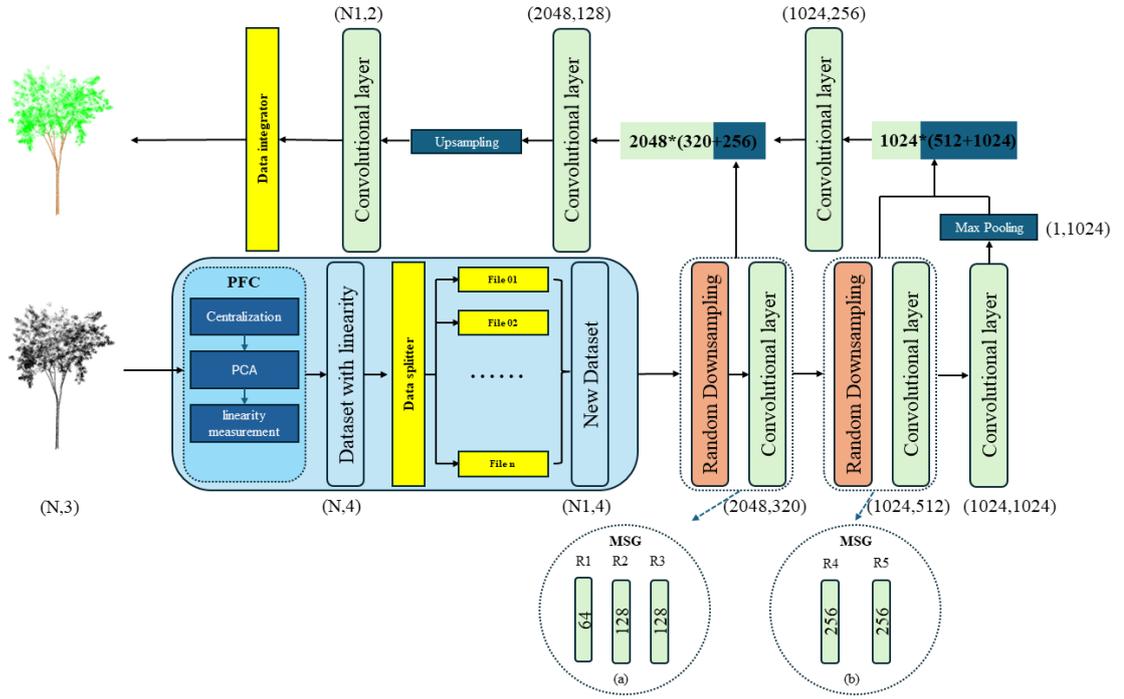

**Figure 4.** WLC-Net structural overview for classifying wood and leaf points in tree point clouds datasets.

(a)、(b)Multi-scale grouping (MSG).

To elevate the wood-leaf classification outcome, WLC-Net enhances classification accuracy, completeness, and speed by incorporating linearity as an inherent feature, refining the input-output framework, and optimizing the centroid sampling technique. The workflow of WLC-Net and experimental design are depicted in Figure 5. Prior Feature Calculation (PFC) module is designed to bolster classification accuracy. Meanwhile, the Splitter and Integrator (SAI) mechanism ensures a complete set of output data. Additionally, the incorporation of a randomized center point selection process serves to curtail processing times. As illustrated in Figure 5, the PFC consists of three components: neighborhood analysis and centralization, principal component analysis (PCA), and linearity measurement. The SAI, on the other hand, is bifurcated into a data splitter and a data output integrator.

In the experiment, leveraging the benchmark classification as a reference, the classification accuracy of different methods was analyzed using three metrics: OA, mIoU, and F1-score. Furthermore, the efficiency of these methods was scrutinized in terms of time cost and time per million points (TPMP), as detailed in the discussion section (Sun

et al., 2021).

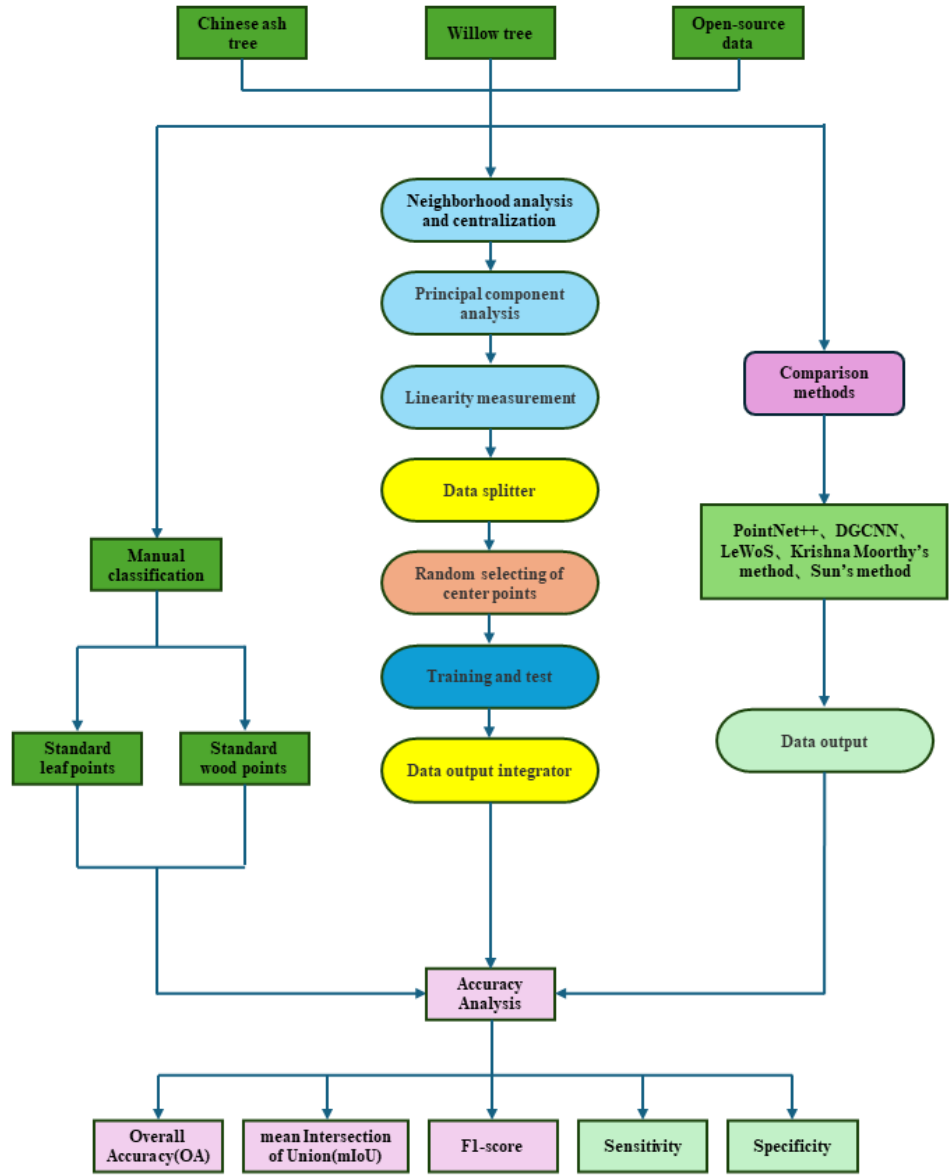

**Figure 5.** The workflow of WLC-Net and experimental design.

### 2.2.1. Prior Feature Calculation (PFC)

A well-established principle in numerous studies posits that integration of prior feature knowledge can markedly improve a model's efficiency in identifying critical information(Sharma et al., 2023). However, this essential feature knowledge is often overlooked in contemporary deep learning techniques, particularly in the domain of wood-leaf classification for tree point cloud (Chen et al., 2021).

To address this, WLC-Net introduced the prior feature.

Prior features are generally categorized into two main categories: radiometric and geometric. Radiometric features, such as the intensity information of tree point clouds, are instrumental in differentiating between wood and leaf points due to their inherent physical disparities. Woody trunks and branches, characterized by their rigidity, contrast with the softer structure and variable positioning of leaves. Furthermore, the wood components, being drier, exhibit a distinct intensity under identical conditions when scanned with the RIEGL VZ-400 scanner, whose laser wavelength coincides with the absorption spectrum of water. However, the utility of intensity values maybe compromised by fluctuations stemming from varying observation angles and environmental conditions, which are prevalent in multi-scan registrations. This variability can impede the application of radiometric features in the wood-leaf classification of registered tree point clouds. In contrast, geometric features are more advantageous for analyzing registered tree point cloud due to the spatial arrangement of wood and leaf points. Wood points exhibit more regularity, reflecting the structured nature of trunk and branches, whereas leaf points are more random and irregular, influenced by their positioning and susceptibility to wind.

Consequently, WLC-Net prioritizes the geometric feature—specifically, the linearity of points—as a pivotal metric for effectively harnessing the geometric information inherent in tree point clouds. The linearity of points, is indicative of the linear alignment within the point set, offering a nuanced and accurate depiction of the structure. The linearity is calculated as following steps:

Step 1: Set the radius of neighborhood.

This process primarily focuses on extracting neighborhood information of each point $P_c$, which is essential for subsequent analysis. The criterion for neighborhood extraction is predicated on the typical diameter of branch, ranging from 0.1m to 0.3m in the middle or lower parts of the tree crown. As depicted in Figure 6, a cylinder symbolizes a branch, with R representing its radius. In this context, $r$ denotes the radius of the neighborhood under consideration. When $r$ is considerably smaller than R, the selected neighborhood approximates a planar configuration, characterized by a scattered distribution of points and a smaller linearity. When $r$ increases to approximately R but remains less than 2R, the neighborhood forms a substantial arc, evidencing a notable linearity. However, when $r$ is marginally greater than 2R, the neighborhood fully encapsulates the branch, exhibiting a pronounced linearity trend. Conversely, if $r$ is considerably larger than 2R, it also incorporates the dispersed leaves in the canopy. Therefore, the selection of $r$ should ideally be no more than 2R and at least equivalent to R. Accordingly, taking into account the morphological characteristics of branches across a majority of tree species, the neighborhood extraction is conducted using a radius $r$ set to 0.15 meters. This results in a point set $M$ comprising all neighbor points and the point of interest $Pc$.

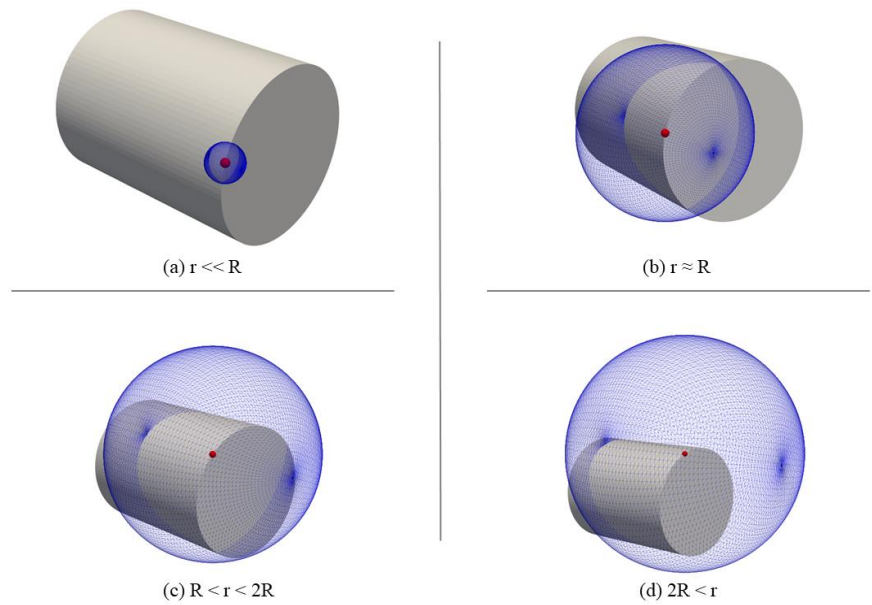

**Figure 6.** The demonstration of the radius selection of the neighborhood. r represents the radius of the neighborhood, and R represents the radius of the branch.

Step 2: Centralize each point of point set $M$.

Centralization is implemented to mitigate the effects of varying scales across different neighborhoods. This process adjusts the raw $X$, $Y$, and $Z$ coordinates of each point in $M$. The centralization of $P_i(x_i, y_i, z_i)$ is fulfilled as follows, $P_i \in M$:

$$P_i'(x_i', y_i', z_i') = (x_i - \bar{x}, y_i - \bar{y}, z_i - \bar{z}), i = 1,2,3, \ldots, n, P_i' \in V \qquad (1)$$

$$\bar{x} = \frac{1}{n}\sum_i^n x_i, \bar{y} = \frac{1}{n}\sum_i^n y_i, \bar{z} = \frac{1}{n}\sum_i^n z_i \qquad (2)$$

Here, $\bar{x}, \bar{y}, \bar{z}$ represent the mean coordinates of the points within $M$. All the centralized points $P_i'$ construct the point set $V$. Centralization is essential for the subsequent process of $V$, like PCA which is invariant to the absolute positions of the points, thereby maintaining the accuracy and reliability of our linearity metric.

Step 3: Perform the PCA analysis of point set $V$.

Now, we get a matrix $V$ which is $3 * n$. To find the principal direction of variance and its corresponding degree, we employ the PCA, a sophisticated dimensionality reduction technique. PCA operates by transforming a dataset with $k$ variables into a new set of $t$ orthogonal variables, known as principal components (Jolliffe and Cadima, 2016). The primary principal component captures the greatest variance within the dataset, and each subsequent component accounts for the maximum remaining variance while adhering to the orthogonality constraint. Following the PCA, the matrix $V$ has been transformed into matrix $X$, a $3 * n$ matrix derived from PCA, and the covariance matrix $\Sigma$ of $X$ is computed as follows:

$$\Sigma = \frac{1}{n-1} X^T X \qquad (3)$$

Step 4: Calculate the linearity of the point.

Because the points in the point set $M$ and the point set $V$ correspond on a one-to-one basis, the linearity for point $P_c$ in $M$ can be calculated as follows:

$$Linearity_{pc} = \frac{\lambda 1 - \lambda 2}{\lambda 1} \qquad (4)$$

The value of linearity ranges from 0 to 1. A linearity value approaching 1 indicates that the point is predominantly aligned along a single axis, indicative a pronounced linear characteristic. In contrast, a linearity value approaching 0 implies a more uniform distribution of the point across multiple axes, signifying a weaker linear characteristic.

Step 5: Calculate the linearity values of point cloud.

Repeat step 2 to step 4 for each point in a tree point cloud and calculate the linearity values for all points. The linearity characteristic of tree ash15 is demonstrated in Figure 7. Linearity values are depicted on a color gradient, where red represents the highest degree of linearity, while blue represents the lowest linearity. It is evident that branches and twigs have the highest linearity, with leaves showing the next highest values, while the trunk exhibits the lowest linearity overall. Furthermore, there is a distinct overlap in the linearity values of leaves and the trunk, whereas the linearity values for branches and twigs are more closely clustered. This observation suggests that the use of linearity as a feature enhances the distinguishability of branches and twigs within the point cloud.

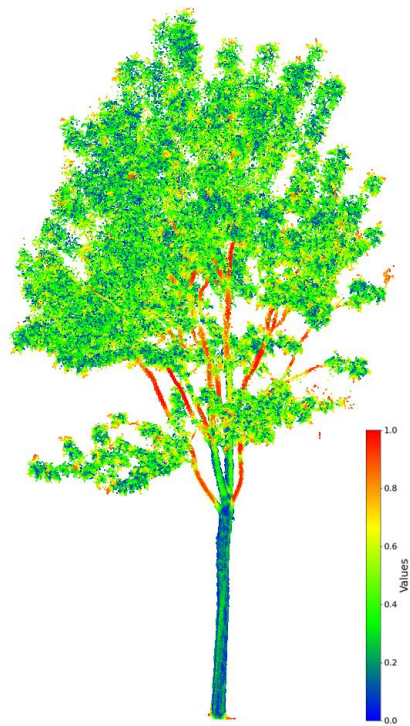

**Figure 7.** Visual representation of linear features (ash15 as an example).

### 2.2.2. Splitter and Integrator

For deep learning methods, more points in the training tree point cloud necessitates more computing resources. Consequently, most deep learning methods are constrained from utilizing excessively large point sets. For instance, datasets generally utilized by algorithms such as PointNet++ and DGCNN typically contain files with exactly 2048 points. And it is also common for deep learning methods to handle point clouds which contained a uniform number of points, typically fewer than 10,000.

For wood-leaf classification tasks, tree point clouds are usually characterized by high density and large number of points, with a single tree in our datasets frequently exceeding 100,000 points. Given the substantial volume of points and the computational resources demanded by deep learning models, WLC-Net employs a strategy to partition the entire tree point cloud into several smaller, more manageable subsets prior to initiating the deep learning analysis.

To achieve this, a splitter function is introduced, designed to adaptively divide tree points into subsets, which can break down a large tree point cloud into several smaller manageable segments, each with a point count as close to a pre-defined maximum $MaxPointNum$ as possible. Taking into account the hardware configuration using in the experiment, the value of $MaxPointNum$ is set to 100,000. The actual number of points contained in each point cloud, $ActualPointNum$, is calculated by:

$$ActualPointNum = \frac{total\ number\ of\ point\ cloud}{MaxPointNum} \qquad (5)$$

Meanwhile, once the WLC-Net model has classified the subsets of points, it is imperative to consolidate these results into a comprehensive dataset. This integration allows for a straightforward comparison of the WLC-Net's outcomes with those of alternative methods. Since the input data have undergone a splitting process and each subset is normalized during the deep learning procedure, it's essential to revert to the original point cloud configuration at the data output stage. Therefore, we perform an inverse normalization on the results of the subsets, then merge the normalized results back together to produce a complete classified tree point cloud.

### 2.2.3. Random selecting of center points

PointNet++ utilizes the Farthest Point Sampling (FPS) method for selecting centroid point, aiming to preserve the original shape of the point cloud as accurately as possible. However, as the number of points increases, so does the burden of calculation which poses challenges for processing dense tree point clouds. The number of points processed now is 100,000, which is much larger than the amount of data processed by PointNet++ model 2048. What's more, upon comparing FPS with other sampling methods like random sampling, density sampling and voxel-based sampling techniques, accuracy differences among these methods are minimal for large-scale tree point cloud, significant variations in computational efficiency are evident.

In WLC-Net, random sampling method is adopted due to its simplicity and computational efficiency(Chen et al., 2015), which is also particularly suited for high-density, large-scale datasets, effectively capturing key geometric and topological features, a quality crucial for complex tree point clouds (Warren and Marz, 2015). Notably, when uniformly sampling same number of points from a point cloud comprising $N$ points, the computational complexity of random sampling is $O(1)$, whereas that of FPS can escalate to $O(N)$. Obviously, the larger the number of points, the more obvious the time advantage of the random sampling method will be compared to FPS. When 2048 centroid points is set in the model, our experiments demonstrated that using random sampling method led to an approximately 45% increase in overall computational efficiency and an improvement in accuracy by about 0.5%.

### 2.2.4. Training and Testing

In this study, three datasets were used, which are Chinese ash, willow, and an open-source dataset. Each of these datasets adhered to a training-test ratio of 2:1. Specifically, the Chinese ash and willow datasets comprised 21 trees, respectively. So, there are 14 trees in the training set and 7 trees in the test set. The open-source dataset initially contained 61 trees. To align with our training-test ratio, we randomly excluded one tree, thereby adopting a training set of 40 trees and a test set of 20 trees.

Furthermore, although the open-source dataset contains 15 different tree species, all of them are tropical trees with relatively two kinds of similar shapes and can be roughly classified into two main categories. Instead of splitting it into two datasets based on category, we trained and tested it as one entire dataset.

Deep learning methods used in the experiment, including the proposed WCL-Net, PointNet++ and DGCNN, are all trained with a learning rate of 0.001, epochs of 60, and a decay rate of 0.5, and conducted under same hardware conditions.

### 2.2.5. Accuracy Analysis

In this study, we utilize three principal metrics to evaluate classification accuracy: OA, mIoU, and the F1-score. OA is a measure of the model's overall classification capability across all categories. It is calculated using the formula:

$$OA = \frac{TP + TN}{TP + TN + FP + FN} \tag{6}$$

In this equation, TP signifies true positives, TN denotes true negatives, FP represents false positives, and FN corresponds to false negatives.

mIoU, a standard metric in deep learning for semantic classification, evaluates the agreement between predicted classifications and actual ground-truth labels(Zhang et al., 2023). For each class, the Intersection over Union (IoU) evaluates the overlap between predicted and actual bounding areas relative to their total combined area. mIoU is the average of all the class specific IoU values. Particularly, in binary classifications, mIoU simplifies to a singular IoU value owing to the evaluation of only two classes. mIoU is formulated as:

$$IoU = \frac{TP}{TP + FP + FN} \tag{7}$$

$$mIoU = \frac{1}{N} \sum_{i=1}^{N} IoU_i \tag{8}$$

However, it's important to note that both OA and mIoU may exhibit bias towards dominant classes. This bias can obscure performance on minority classes, a critical consideration in tree points cloud data where disparities between wood and leaf points can reach ratios as high as 1:10. Relying solely on OA and mIoU could lead to performance indicators that inadvertently favor these dominant classes.

For a comprehensive evaluation of our model, we primarily focus on the F1-score, while also recognizing the importance of the Kappa Coefficient, another well-regarded metric. The Kappa Coefficient is particularly useful as it accounts for chance agreement, it highlights how the model's performance exceeds mere random chance, a consideration crucial for multi-class problems where random predictions can significantly skew results(Chicco et al., 2021).

However, it's important to note that in binary classification scenarios, particularly with highly imbalanced sample sizes, the Kappa Coefficient might sometimes present an overly optimistic view of the model's effectiveness. In such cases, we turn to the F1-score for a more accurate assessment. The F1-score is particularly adept at emphasizing the performance of the positive class, which is often in the minority(Guo et al., 2021). This makes it an ideal metric for distinguishing between categories like wood and leaf points in tree point clouds. The F1-score combines precision and recall in a balanced manner, making it invaluable in scenarios with significant data imbalances. It is calculated as follows:

$$Precision = \frac{TP}{TP + FP} \tag{9}$$

$$Recall = \frac{TP}{TP + FN} \tag{10}$$

$$F1 = 2 \times \frac{Precision \times Recall}{Precision + Recall} \tag{11}$$

In conclusion, by utilizing a combination of evaluation metrics—OA, mIoU, and the F1-score—we ensure a thorough and insightful analysis of the model's performance. This diversified strategy is essential in dealing with tree point clouds, where the disparity between classes presents a considerable challenge. This rigorous evaluation framework aims to confirm the model's accuracy and its proficiency in distinguishing complex tree structures.

Besides, considering the different characteristics, and the open-source dataset are evaluated in the reference article using sensitivity and specificity. For convenience, the calculations of sensitivity and specificity are listed as follows. Sensitivity also known as the true positive rate, measures the proportion of actual positives correctly identified by the model. In contrast, specificity, or the true negative rate, assesses the proportion of actual negatives accurately identified.

$$Sensitivity = \frac{TP}{TP + FN} \tag{12}$$

$$Specificity = \frac{TN}{TN + FP} \tag{13}$$

## 3. Results

In our experiment, a consistent training-to-test ratio of 14:7 was applied across both Chinese ash trees and willows. As previously mentioned, the manual classification results of all trees served as the standard benchmarks. The trees allocated to the testing phase—comprising 7 Chinese ash trees and 7 willow trees—were pivotal for our analytical process.

Furthermore, to verify the applicability of WLC-Net, the open-source dataset was also used, which includes a diverse range of tree species. Considering the dataset's composition of multiple tree species with a limited number per type, each species, we elected to test it in its entirety. To maintain the consistent training-test ratio of 2:1, one tree was randomly excluded from the dataset, resulting in 60 trees remaining. Consequently, 40 tree point clouds were designated for the training set, while 20 tree point clouds were allocated to the testing set.

### 3.1. WLC-Net classification results

The performance of WLC-Net was evaluated across three key dimensions: visual inspection, accuracy, and efficiency. As previously mentioned, the testing phase included 7 tree point clouds each for Chinese ash and willow trees. The classification results for these 14 trees are presented in Figure 8 and 9, with wood points depicted in brown and leaf points in green. It is notable that misclassifications of leaf points on the main trunks and primary branches are rare, indicating the superior performance of WLC-Net in distinguishing trunk and branch points. The classification results for the 20 test samples from the open-source dataset are extensively detailed in Figure 10. It is shown that WLC-Net maintains robust performance on the shared dataset, even when scanning parameters and specific tree characteristics are not known. The trio of visual presentations corroborates the high reliability and broad applicability of WLC-Net across diverse tree point clouds.

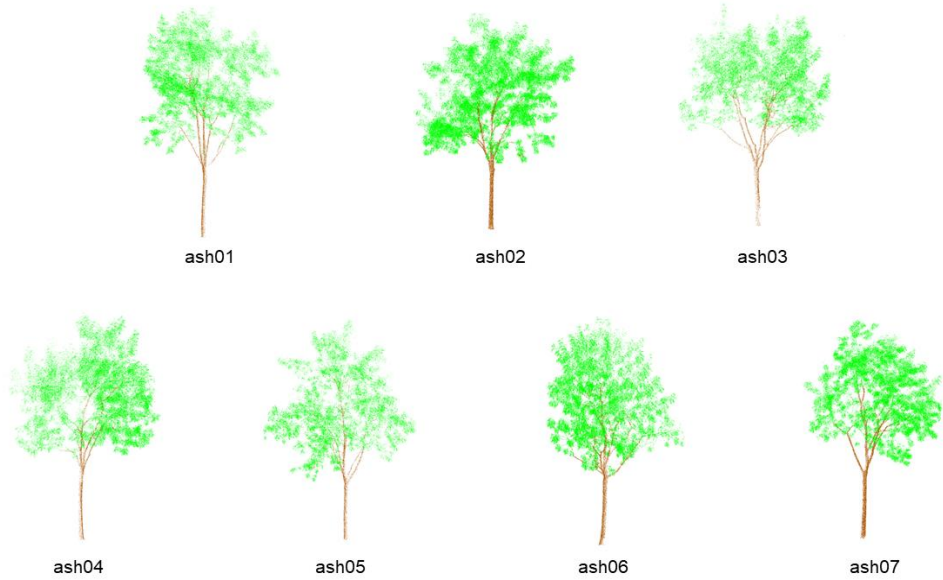

**Figure 8.** The demonstration of 7 Chinese ash classification results. Green: leaf points; brown: wood points.

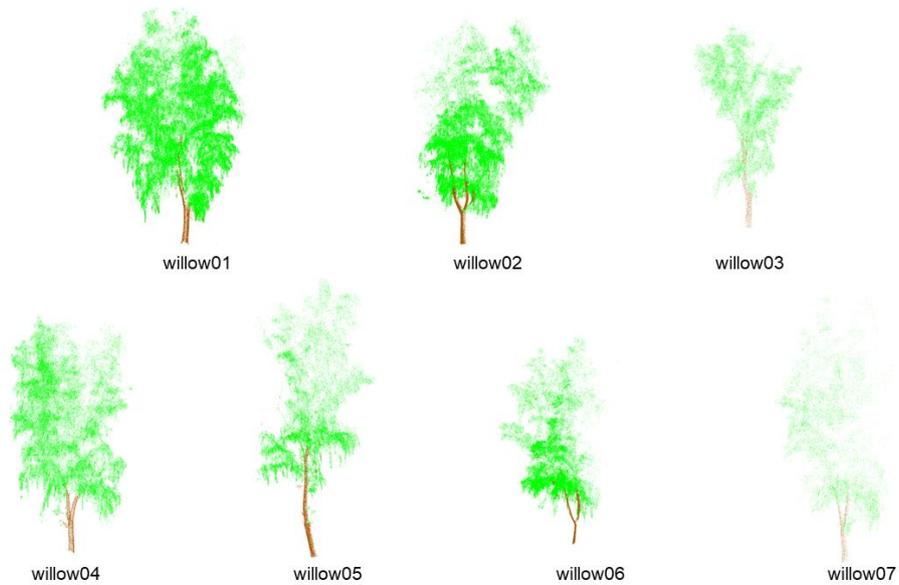

**Figure 9.** The demonstration of 7 willow trees classification results. Green: leaf points; brown: wood points.

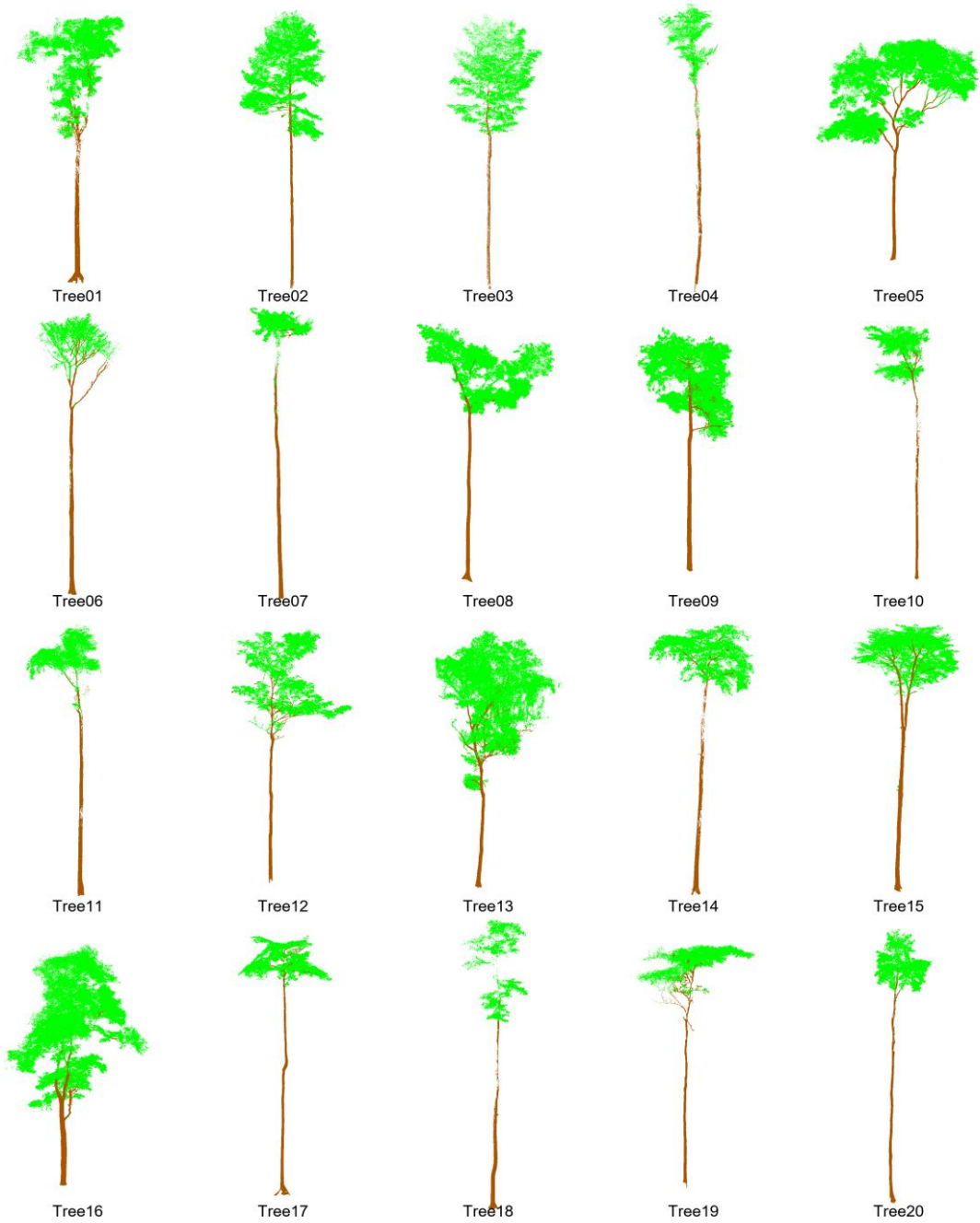

**Figure 10.** The demonstration of 20 open-source data classification results. Green: leaf points; brown: wood points.

The accuracy evaluation of WLC-Net was grounded in three key metrics: OA, mIoU and F1-score. Comprehensive results for three datasets are presented in Table 2. As listed in Table 2, for our two datasets, OA values range from 0.9554 to 0.9899, with the majority of trees surpassing the 0.96 threshold. The average OA for

Chinese ash trees is 0.9778, while for willows, it stands at 0.9712. Notably, mIoU values generally align with OA, when the proportion of contrasting samples varies markedly, suggesting a positive association. However, the F1-scores exhibit a wider variance, spanning from 0.5804 to 0.9216, with no discernible pattern in relation to OA or mIoU.

In comparison with the above visual demonstrations, it is apparent that the trees recording lower F1-scores always have obvious main trunk and elusive canopy interior wood points. These trees often exhibit suboptimal classification upon visual inspection, yet are challenging to identify using OA and mIoU alone. Based on the datasets of Chinese ash and willow, the correlation diagrams for OA against mIoU and F1-score are depicted in Figure 11. The $R^2$ for OA with mIoU and OA with F1-score are 0.9917 and 0.6235, respectively. Evidently, OA correlates strongly with mIoU, whereas the relationship with F1-score is less pronounced. This highlights the importance of the F1-score as a stringent and precise metric for evaluating wood-leaf classification.

For the open-source dataset, a comprehensive evaluation was conducted, encompassing not only OA, mIoU and F1-score, but also sensitivity and specificity. As reported, LeWoS exhibited an OA of 0.91 ± 0.03, a sensitivity of 0.92 ± 0.04 and a specificity of 0.89 ± 0.06. In our experimental evaluation, we utilized LeWoS to assess the subset of 20 tree point clouds extracted from the open-source dataset. This analysis yielded an OA of 0.9109, a sensitivity of 0.9539, and a specificity of 0.8256. These discrepancies are likely attributed to the use 20 trees from the dataset's total of 61 trees. Furthermore, to facilitate a direct comparison between the WLC-Net and LeWoS methods, we also evaluated the performance of WLC-Net on the total open-source dataset using sensitivity and specificity. This analysis achieved a sensitivity of 0.9878 and a specificity of 0.8565. Obviously, this highlights the robust capability of WLC-Net to distinguish between wood and leaves, ensuring high precision in identifying wood points while maintaining a lower rate of incorrectly labeling leaf points as wood points.

**Table 2.** The accuracy analysis of 34 trees classification results.

| Tree Species | Numerical Identifier | OA | mIoU | F1-score |
|---|---|---|---|---|
| Chinese ash | 01 | 0.9860 | 0.9827 | 0.9216 |
| | 02 | 0.9787 | 0.9774 | 0.8400 |
| | 03 | 0.9755 | 0.9735 | 0.8650 |
| | 04 | 0.9788 | 0.9775 | 0.8393 |
| | 05 | 0.9774 | 0.9758 | 0.8561 |
| | 06 | 0.9652 | 0.9624 | 0.8091 |
| | 07 | 0.9827 | 0.9811 | 0.9084 |
| Avg. | | 0.9778 | 0.9761 | 0.8628 |
| Willow | 01 | 0.9554 | 0.9540 | 0.5804 |
| | 02 | 0.9707 | 0.9679 | 0.8565 |
| | 03 | 0.9668 | 0.9653 | 0.7191 |
| | 04 | 0.9899 | 0.9896 | 0.8618 |
| | 05 | 0.9770 | 0.9742 | 0.9045 |
| | 06 | 0.9667 | 0.9639 | 0.8251 |
| | 07 | 0.9717 | 0.9699 | 0.8094 |
| Avg. | | 0.9712 | 0.9693 | 0.7938 |
| Open-source data | 01 | 0.9554 | 0.9646 | 0.9317 |
| | 02 | 0.9485 | 0.9719 | 0.9453 |
| | 03 | 0.9650 | 0.9617 | 0.9805 |
| | 04 | 0.9408 | 0.9197 | 0.8988 |
| | 05 | 0.9645 | 0.9600 | 0.9796 |
| | 06 | 0.9289 | 0.7187 | 0.8364 |
| | 07 | 0.9711 | 0.9511 | 0.9749 |
| | 08 | 0.9578 | 0.9380 | 0.9680 |
| | 09 | 0.9139 | 0.8918 | 0.9428 |
| | 10 | 0.9597 | 0.9422 | 0.9703 |
| | 11 | 0.9598 | 0.8743 | 0.9329 |
| | 12 | 0.9158 | 0.9009 | 0.9479 |
| | 13 | 0.9495 | 0.9349 | 0.9664 |
| | 14 | 0.9425 | 0.9128 | 0.9544 |

| | 15 | 0.9140 | 0.8365 | 0.9110 |
|---|---|---|---|---|
| | 16 | 0.9717 | 0.9676 | 0.9835 |
| | 17 | 0.9647 | 0.9506 | 0.9747 |
| | 18 | 0.9851 | 0.8991 | 0.9468 |
| | 19 | 0.9402 | 0.9110 | 0.9534 |
| | 20 | 0.9776 | 0.9565 | 0.9777 |
| Avg. | | 0.9513 | 0.9182 | 0.9489 |

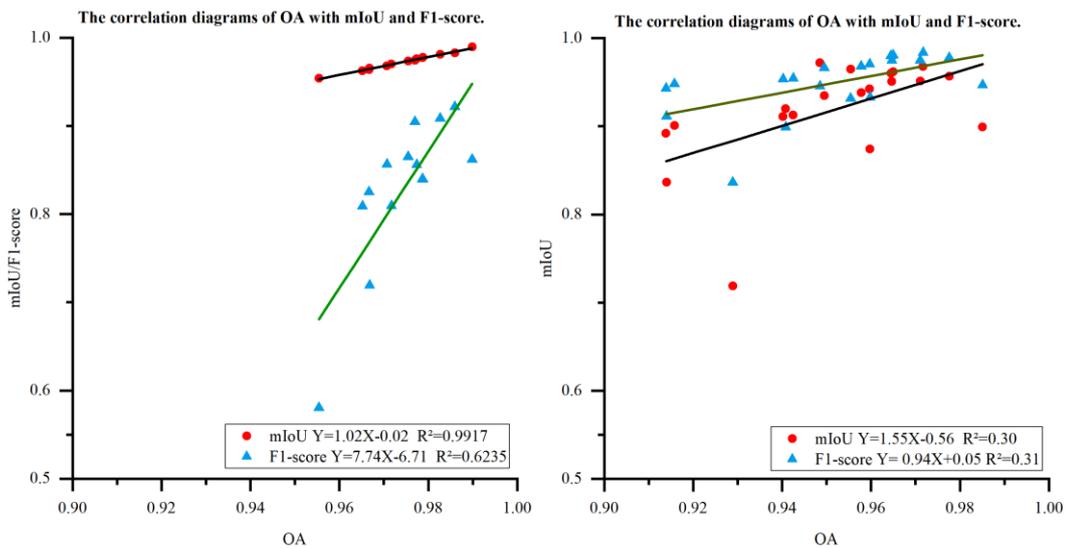

**Figure 11.** The correlation diagrams of OA with mIoU and F1-score based on three datasets. The graph on the left is made up of Chines ash and Willow datasets, and the one on the right is made up of Open-source datasets.

The efficiency evaluation is primarily based on the processing time and TPMP of individual tree point cloud. Comprehensive results of three datasets are presented in Table 3. As detailed in Table 3, the total points of each data varies significantly, ranging from 21608 to 8269426. This disparity directly influences the processing time required for each dataset, which varies from a mere 4.39 seconds to a more substantial 923.04 seconds. The correlation between the total points and the processing time is illustrated in Figure 12, demonstrating a predominantly linear relationship with an $R^2$ value of 0.9484, indicating a strong correlation between the total points and the processing time. Additionally, the TPMP ranges from 62.09 to 203.26 seconds, with an average of 102.74 seconds.

The majority of the datasets exhibit a TPMP within the range of 80 and 120 seconds. Notably, the highest TPMP value of 203.26 seconds is associated with the tree point cloud with the fewest points, numbering 21608. This suggests that a considerable portion of the processing time is consumed by operations that are not directly proportional to the number of points processed, such as model invocation and file I/O operations.

Table 3. The time cost of 34 tree point cloud in three testing datasets.

| Tree Species | Numerical Identifier | Total Points | Time cost(s) | TPMP(s) |
| --- | --- | --- | --- | --- |
| Chinese ash | 01 | 107024 | 10.90 | 101.81 |
| | 02 | 228642 | 26.42 | 115.57 |
| | 03 | 96460 | 10.22 | 105.99 |
| | 04 | 127539 | 15.55 | 121.94 |
| | 05 | 83287 | 8.90 | 106.91 |
| | 06 | 175362 | 16.70 | 95.25 |
| | 07 | 182112 | 17.18 | 94.36 |
| Willow | 01 | 430625 | 41.52 | 96.42 |
| | 02 | 307784 | 29.44 | 95.65 |
| | 03 | 43697 | 6.49 | 148.48 |
| | 04 | 161474 | 17.13 | 106.07 |
| | 05 | 92763 | 10.83 | 116.77 |
| | 06 | 140854 | 16.72 | 118.70 |
| | 07 | 21608 | 4.39 | 203.26 |
| Open-source dataset | 01 | 2584586 | 199.62 | 77.23 |
| | 02 | 1118126 | 119.79 | 107.13 |
| | 03 | 279259 | 30.87 | 110.54 |
| | 04 | 140889 | 14.11 | 100.15 |
| | 05 | 5887978 | 662.75 | 112.56 |
| | 06 | 1626003 | 153.25 | 94.25 |
| | 07 | 4082518 | 288 | 70.54 |
| | 08 | 4208202 | 345.3 | 82.05 |

|     |         |        |        |
| --- | ------- | ------ | ------ |
| 09  | 5472253 | 541.8  | 99.01  |
| 10  | 814805  | 67.14  | 82.40  |
| 11  | 1545318 | 157.54 | 101.95 |
| 12  | 877566  | 86.8   | 98.91  |
| 13  | 3767575 | 345.6  | 91.73  |
| 14  | 1361736 | 106.5  | 78.21  |
| 15  | 2735498 | 169.76 | 62.06  |
| 16  | 8269482 | 923.04 | 111.62 |
| 17  | 1859161 | 158.55 | 85.28  |
| 18  | 2227416 | 339.32 | 152.34 |
| 19  | 2608974 | 220.59 | 84.55  |
| 20  | 2432919 | 154.7  | 63.59  |
| Avg.|         | 1649985 | 156.40 | 102.74 |

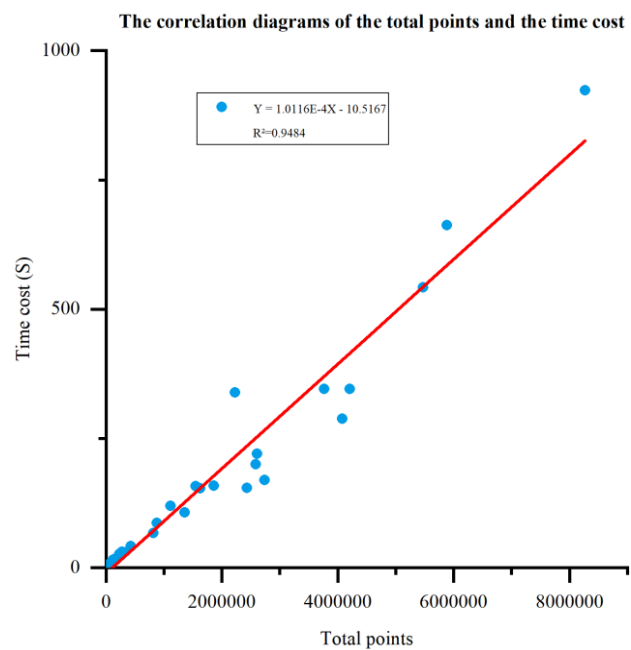

**Figure 12.** The correlation diagrams of the total points and the time cost.

### 3.2. Accuracy analysis

To thoroughly assess the performance of WLC-Net, further experiments were undertaken, comparing it against five alternative methods: PointNet++, DGCNN, Krishna Moorthy's method, LeWoS, and Sun's method. PointNet++ and DGCNN are prominent deep learning techniques, Krishna Moorthy's method is a distinguished machine learning strategy, and both LeWoS and Sun's method are recognized for their effectiveness as traditional approaches. Table 4 provides a quantitative comparison of WLC-Net with these five comparator methods. Based on three metrics, WLC-Net 's wood-leaf classification accuracy outperformed the other five methods in a comprehensive manner.

For Chinese ash tree point clouds, WLC-Net and LeWoS demonstrated closely competitive results, markedly outperforming the other four methods. WLC-Net improved upon LeWoS by 0.09% in OA, 0.08% in mIoU, and 1.4% in F1-score. Among the latter four methods, PointNet++ exhibits the best performance, with Sun's method and DGCNN lagging behind. Specifically, for ash tree processing, WLC-Net, LeWoS, and PointNet++ exhibited OA values above 90%, significantly higher than the OA values of the remaining three methods, which were all below 90%.

For willow tree point clouds, the results from WLC-Net and LeWoS are still quite close, markedly outperforming the other four methods. WLC-Net improved upon LeWoS by 1.37% in OA, 1.51% in mIoU, and 3.66% in F1-score. Compared to processing Chinese ash tree point clouds, the leading advantage of WLC-Net over LeWoS has increased slightly. The results of PointNet++ and DGCNN fall into an excellent second tier, and their metrics are very close to each other. Their OA values approximate 92%, which is a notable improvement over Krishna Moorthy's method and Sun's method, both of which have OA values below 90%. In contrast, both WLC-Net and LeWoS achieve OA values in excess of 95%, clearly delineating their better performance. Additionally, the F1-score for WLC-Net and LeWoS exceed 75%, surpassing the other four methods significantly, which are around 40%. This disparity highlights the distinct advantages of WLC-Net and LeWoS in the classification of willow tree point clouds.

For the open-source dataset, which is devoid of intensity data, Sun's method was excluded from testing. Consequently, the remaining five methods were evaluated. Among these, only the Krishna Moorthy's method had an OA below 90%, while the other four methods all exceeded this threshold. Notably, WLC-Net and PointNet++ demonstrated significantly higher OA compared to LeWoS and DGCNN. Specifically, WLC-Net and PointNet++

achieved OA values above 93%, whereas LeWoS and DGCNN had OA values around 91%. In particular, WLC-Net improved upon PointNet++ by 2.04% in OA, 3.23% in mIoU, and 4.68% in F1-score. And all the five methods achieved F1-score accuracies above 80%, with WLC-Net notably reaching an F1-score accuracy of over 90%.

Table 4. Wood-leaf classification results comparison of six methods.

| Tree species | Related methods | OA | mIoU | F1-score |
|---|---|---|---|---|
| Chinese ash | PointNet++ | 0.9590 | 0.9566 | 0.7254 |
| | DGCNN | 0.8873 | 0.8819 | 0.4249 |
| | Krishna Moorthy's method | 0.8965 | 0.8897 | 0.5461 |
| | LeWos | 0.9769 | 0.9753 | 0.8488 |
| | Sun's method | 0.8727 | 0.8662 | 0.4834 |
| | WLC-Net | 0.9778 | 0.9761 | 0.8628 |
| Willow | PointNet++ | 0.9226 | 0.9192 | 0.4986 |
| | DGCNN | 0.9146 | 0.9121 | 0.3785 |
| | Krishna Moorthy's method | 0.8642 | 0.8552 | 0.4741 |
| | LeWos | 0.9575 | 0.9542 | 0.7572 |
| | Sun's method | 0.7836 | 0.7688 | 0.3794 |
| | WLC-Net | 0.9712 | 0.9693 | 0.7938 |
| Open-source Data | PointNet++ | 0.9304 | 0.8818 | 0.8551 |
| | DGCNN | 0.9168 | 0.8661 | 0.8268 |
| | Krishna Moorthy's method | 0.8464 | 0.6317 | 0.8396 |
| | LeWos | 0.9109 | 0.8523 | 0.8372 |
| | Sun's method | - | - | - |
| | WLC-Net | 0.9508 | 0.9141 | 0.9019 |

### 3.3. Efficiency analysis

To better evaluate WLC-Net, we also compared its processing efficiency against other methods. However, since DGCNN, Krishna Moorthy's method, and LeWoS did not provide explicit timing data, we resorted to manually estimatingthe processing time based on ash01 with around 100000 points. Approximately, DGCNN required about 10 seconds, Krishna Moorthy's method took around 150 seconds, and LeWoS used about 60 seconds. As mentioned above，Sun's method, unable to handle point cloud data without intensity values, is absent from the efficiency test on the open-source dataset. Meanwhile, PointNet++ encountered hardware limitations, notably insufficient GPU memory, leading to program interruptions. In terms of processing speed, Sun's method is remarkably fast, completing most tree point cloud in one second. WLC-Net and PointNet++ are more analogous, they are both deep learning approaches. According to our test, for tree point cloud with around 100,000 points or fewer, WLC-Net achieves an efficiency improvement of about 41% over PointNet++. As the number of points increases, PointNet++ experiences an exponential rise in processing time, which can eventually lead to a GPU memory exhaustion. In contrast, WLC-Net demonstrates a linear increase in processing time with the number of points increases, reflecting its more adaptable performance in handling larger datasets.

## 4. Discussion

According to the preceding information, WLC-Net exhibits the best overall performance in wood-leaf classification accuracy across three datasets. The incorporation of linearity as a feature-level discriminator has significantly enhanced the model's performance, achieving an improvement of approximately 2.93% in OA, 3.4% in mIoU, and 15.98% in F1-score across three datasets when compared to the original PointNet++ model. This enhancement highlights the effectiveness of using linearity as a prior feature to augment the performance of the well-established PointNet++ model. Additionally, we addressed the original PointNet++ model's limitation regarding the complete input and output of the entire tree point cloud, a challenge that, despite the calculation of OA and mIoU, the model could not previously overcome.

However, the implementation of linearity as a feature encounter specific challenge. Ideally, branches, twigs,

and trunks should exhibit higher linearity values, yet in practice, branches and twigs show the highest linearity, and leaf points often surpass even the linearity values of trunk points. The calculation of linearity is closely tied to the scale of the selected neighborhood, which suggests that while branches exhibit greater linearity than leaves, leaves in turn show greater linearity than the trunk. This distinction, particularly between trunk and branch parts, is quite pronounced and may impact the accuracy of wood-leaf classification.

Moreover, we observed some notable trends. For instance, all methods generally perform well on open-source data, particularly the three deep learning methods, which show excellent overall performance. However, in the case of Chinese ash and willow trees, apart from WLC-Net, the performance of various algorithms varies, especially since some methods show decent OA values but poor F1-score performance. In our analysis of the two datasets, we selected several trees with F1-scores below the average for closer observation. Compared to trees with higher F1-scores, these trees tend to have denser canopies, and most also have a lower ratio of wood points and leaf points than the average. As demonstrated in Figure 13, we present three such trees along with their corresponding standard data, F1-scores, and the ratio of wood points and leaf points. For willow01 and ash02, the unrecognized fine branches were mostly located in areas obscured by the canopy. However, intriguingly, the most significant omissions in ash04 were on the left side where the fine branches were not heavily covered by the canopy. Therefore, we speculate that the performance of these methods may be related to the scanning parameters, the tree morphology and the ratio of wood points and leaf points in the point cloud.

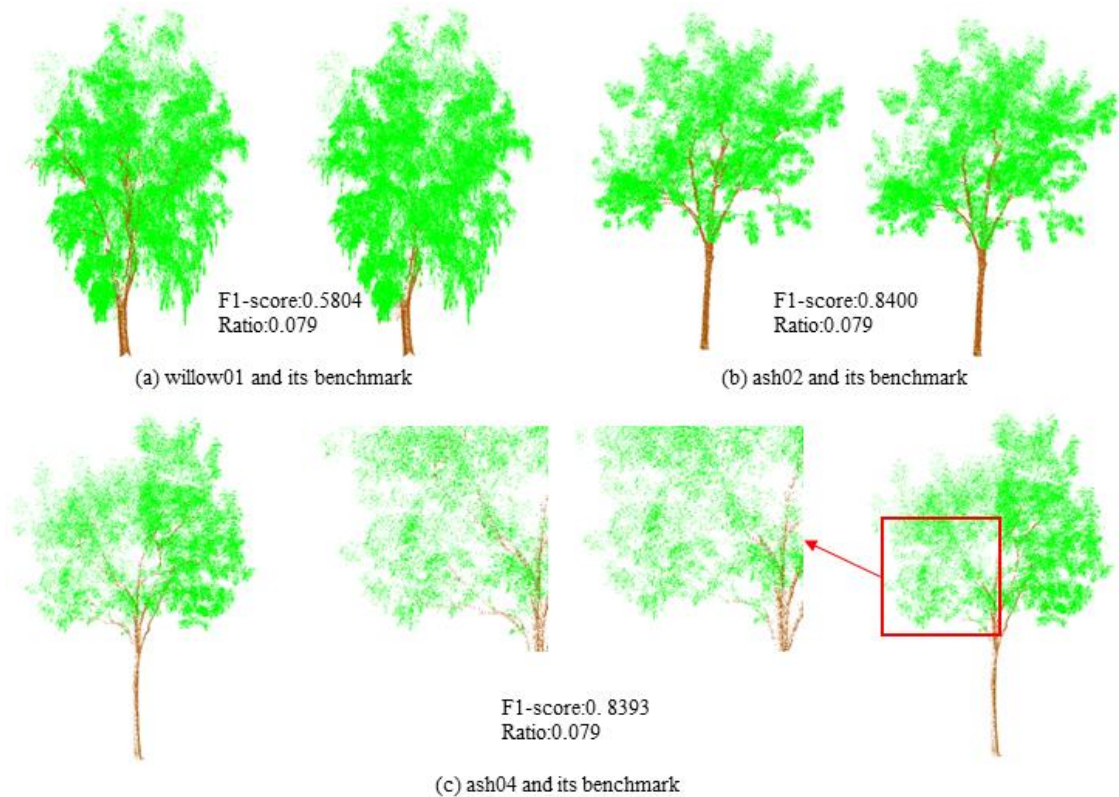

**Figure 13.** Comparison of benchmark and WLC-Net results for willow01, ash02 and ash04. Green: leaf points; brown: wood points.

To verify our hypotheses and further assess the performance of WLC-Net on different datasets, we tested a single-scan willow dataset from 2016, which was collected with a scanning resolution of 0.02 degrees. As shown in Figure 14, the classification accuracy for the eight trees within the dataset is quite satisfactory, suggesting that scanning parameters do not markedly impact the accuracy for this specific dataset. Although the OA for all trees is high, there is a notable variation in the F1-scores. Especially noteworthy is the tree labeled Single-scan04, which has an F1-score of 0.6181, significantly below the average and indicating the poorest classification among the sampled trees. This discrepancy is visually confirmed in Figure 14 and highlights the F1-score's utility in evaluating wood-leaf classification. However, upon closer examination of Figure 14, it is evident that that Single-scan04 is also the most heavily obscured by the canopy. Additionally, the binary classification problem in F1-score calculations is inherently influenced by the ratio of sample points, making it difficult to conclusively determine whether the low F1-scores are due to canopy obstruction or a low ratio of wood to leaf points.

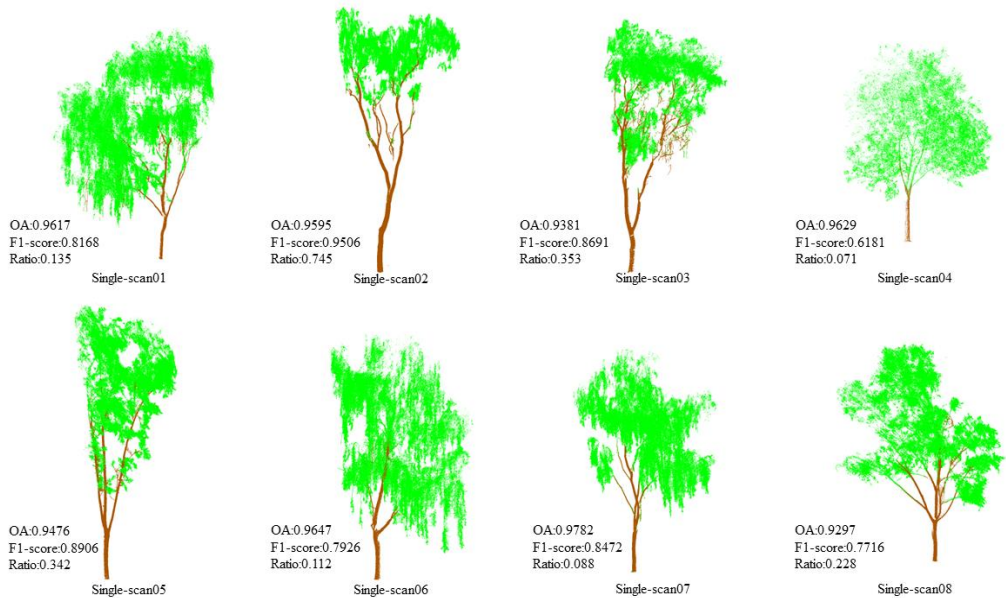

**Figure 14.** WLC-Net classification results of 8 single-scanned willow trees. Green: leaf points; brown: wood points.

To assess the influence of the two key factors, —canopy obstruction and the ratio of wood to leaf points— we turned to the open-source dataset featuring trees with heights generally exceed 30 meters. These trees exhibit a low vertical distribution ratio of the canopy, indicating that the canopy's vertical span, from the lowest to the highest points, constitutes a small fraction of the overall tree height. Consequently, the ratio of wood to leaf points is significantly high, and the points are less likely to be obscured by the canopy. However, as illustrated by tree06 in Figure 15, despite the favorable conditions of a high wood-to-leaf point ratio and minimal obstruction, the classification performance remained suboptimal. Only the trunk and primary branches were accurately identified, with finer branches receiving minimal classification. This outcome may be attributed to the substantial morphological differences between the trunk and branches, particularly in terms of linearity. It is conceivable that the dataset's predominance of trunk points over branch points introduces a bias during the model's training phase. With the majority of wood points originating from the trunk, the model may predominantly learn to recognize the trunk, leading to inadequate recognition and training on finer branches. This dataset imbalance likely plays a role in the reduced effectiveness in recognizing branches, even when there is no canopy obstruction to hinder the process.

In datasets like Chinese ash and Willow, the observed imbalance is significantly influenced by both canopy coverage and the ratio of wood to leaf points, leading us to previous conclusion that these two factors significantly affect classification results. To address this, we employed CloudCompare to cropping the trunk portions from the open-source dataset and retrained the model. Our revised strategy concentrated on training the model specifically on the branch components. The result of this refined training is shown in 15(b).

Given the modifications made to the original data by cropping the trunk, it is not reasonable to compare accuracy directly with the unaltered dataset. Thus, we visually assessed the classification results. Obviously, the comparison between Figure 15(a) and 15(b) shows that training on the dataset after cropping the trunks yields better classification results.

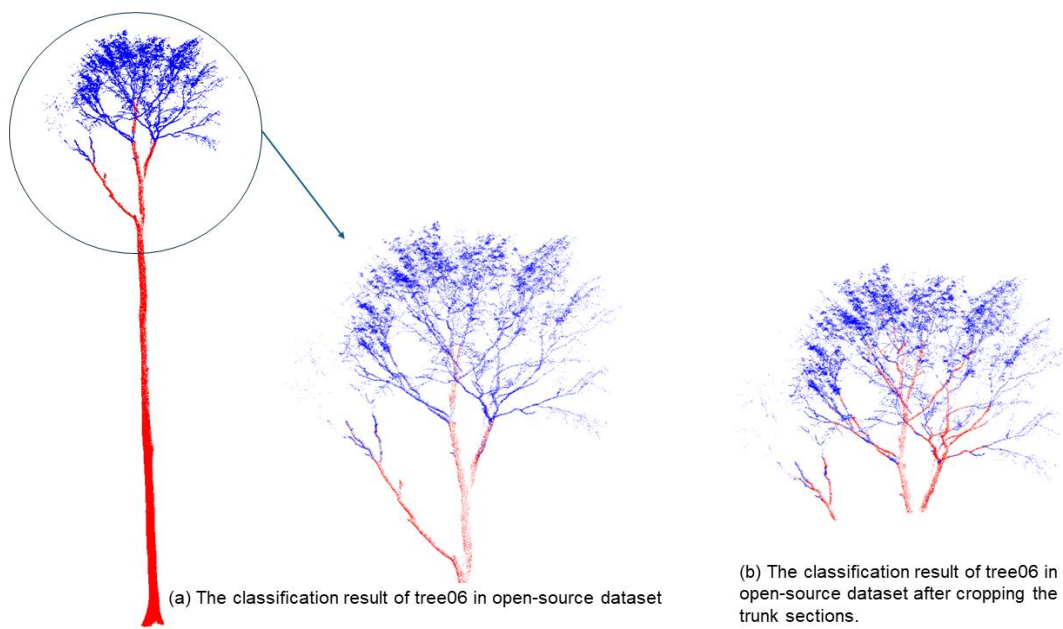

**Figure 15.** Comparison of classification results for Tree06 before and after trunk cropping.

Therefore, for wood-leaf classification, a tri-class scheme, including trunk, branches, and leaves, may be more appropriate. This methodology has the potential to reduce the inconsistencies observed in the classification of trunk and branch segments. Implementing a tri-class system could augment the wood-leaf classification accuracy within the realm of deep learning. Consequently, this enhancement could bolster the model's versatility and robustness.

In terms of efficiency, there is a notable variation in processing time across the different methods utilized. Sun's approach, which classifies tree point cloud directly based on its intrinsic characteristics, without the need for sample collection or model training, and is implemented in C++. This method is the fastest, capable of processing individual tree point cloud within approximately one second. The LeWoS method, developed in Matlab, is also without the need for samples or training and is relatively slower. Krishna Moorthy's method, a machine learning-based approach executed in Python, exhibits longer processing times when compared to LeWoS. Deep learning methodologies necessitate a training phase, which involves an initial investment of time to compile a dataset and train the model. However, once the model is trained, these methods generally outpace LeWoS in terms of processing time.

Among the deep learning approaches, WLC-Net stands out for its superior speed, significantly outperforming PointNet++. This enhancement can be ascribed to two primary factors: the employment of a random sampling strategy instead of the more traditional farthest point sampling during centroid selection; and the segmentation of extensive point cloud data into more manageable chunks for processing, thereby conserving computational resources. As shown in Table 5, the time efficiency for the Chinese ash and Willow datasets is detailed, with a distinction made between tree point clouds that required splitting and that which did not. For the data requiring split, PointNet++ takes a considerable amount of time due to hardware limitations. Specifically, when the point cloud data surpasses a certain threshold—approximately 120000 points under our testing conditions—the processing time escalates sharply, potentially leading to GPU memory exhaustion and program failure. In these scenarios, the efficiency gains of WLC-Net are predominantly attributable to its data splitting module. For datasets that do not require splitting,

WLC-Net operates approximately 41% faster than PointNet++, with this improvement largely due to the centroid sampling technique it employs.

Table 5. Efficiency comparison against WLC-Net and PointNet++.

|  | Tree | WLC-Net | PointNet++ |
|---|---|---|---|
| Split | ash02 | 26.42 | 454.072 |
|  | ash04 | 15.55 | 48.456 |
|  | ash06 | 16.70 | 216.064 |
|  | ash07 | 17.18 | 235.008 |
|  | willow02 | 29.44 | 726.18 |
|  | willow04 | 17.13 | 174.99 |
|  | willow06 | 16.72 | 121.88 |
| Avg. |  | 19.88 | 282.38 |
| Intact | ash01 | 10.90 | 14.272 |
|  | ash03 | 10.22 | 13.872 |
|  | ash05 | 8.90 | 13.072 |
|  | willow03 | 6.49 | 10.208 |
|  | willow05 | 10.83 | 13.344 |
|  | willow07 | 4.39 | 8.376 |
| Avg. |  | 8.62 | 12.19 |

## 5. Conclusions

This research presents WLC-Net, a pioneering method for the automated wood-leaf classification within tree point cloud data. By adapting the PointNet++ architecture, we integrated linearity as a potent distinguishing feature, which significantly aids in precise identification of branches and leaves. Quantitative assessments, anchored on manually classified benchmarks, validate the F1-score's reliability in gauging model efficiency. WLC-Net demonstrates exceptional performance in terms of accuracy, computational efficiency, and robustness, especially when dealing with variations in data quality and tree species. Future research should prioritize refining the linearity feature and investigating multi-class classification techniques to navigate the challenges of classifying intricate and obscured canopy structures. Through these endeavors, we aspire to elevate the model's capabilities and widen its relevance to diverse tree point cloud datasets.